%% file: paper-acc-2016-00-01-main.tex
\documentclass[letterpaper, 10pt, conference]{ieeeconf}  
\IEEEoverridecommandlockouts                              
\usepackage[dvips]{graphics} 
\usepackage{epsfig} 
\usepackage{mathptmx} 
\usepackage{mathtools} 
\usepackage{times} 
\usepackage{amsmath} 
\usepackage{amssymb}  
\usepackage{txfonts}  

\usepackage{algorithm}
\usepackage{algorithmic}
\DeclareMathOperator*{\argmin}{arg\,min}
\DeclareMathOperator*{\argmax}{arg\,max}
\usepackage{subcaption}
\usepackage{gnuplot-lua-tikz}
\usepackage{tikz}
\newtheorem{theorem}{Theorem}

\input{paper-acc-2016-01-01-title.tex}

\begin{document}

\maketitle
\thispagestyle{empty}
\pagestyle{empty}

\input{paper-acc-2016-02-01-abst.tex}

\input{paper-acc-2016-03-01-intro.tex}

\input{paper-acc-2016-05-00-em.tex}

\input{paper-acc-2016-05-01-em.tex}

\input{paper-acc-2016-06-00-dqaem.tex}

\input{paper-acc-2016-06-01-dqaem.tex}

\input{paper-acc-2016-06-02-dqaem.tex}

\input{paper-acc-2016-07-00-result.tex}

\input{paper-acc-2016-07-01-result.tex}

\input{paper-acc-2016-07-02-result.tex}

\input{paper-acc-2016-07-03-result.tex}

\input{paper-acc-2016-07-04-result.tex}

\input{paper-acc-2016-07-05-result.tex}

\input{paper-acc-2016-08-02-conc.tex}


\bibliographystyle{IEEEtran}
\bibliography{IEEEabrv,paper-acc-2016-99-01-bib}



\end{document}

%% file: paper-acc-2016-01-01-title.tex
\title{\LARGE \bf
Relaxation of the EM Algorithm via Quantum Annealing*
}

\author{Hideyuki Miyahara and Koji Tsumura
\thanks{*This work was supported in part by
  Grant-in-Aid for Scientific Research (B) (25289127), Japan
  Society for the Promotion of Science.}
\thanks{H. Miyahara and K. Tsumura are with
 the Department of Information Physics and Computing, the Graduate School of Information Science and Technology, The University of Tokyo, Japan
  {\tt\small hideyuki\_miyahara@ipc.i.u-tokyo.ac.jp}}%
}

%% file: paper-acc-2016-02-01-abst.tex
\begin{abstract}
The EM algorithm is a novel numerical method to obtain maximum likelihood estimates and is often used for practical calculations.
However, many of maximum likelihood estimation problems are nonconvex, and it is known that the EM algorithm fails to give the optimal estimate by being trapped by local optima.
In order to deal with this difficulty, we propose a deterministic quantum annealing EM algorithm by introducing the mathematical mechanism of quantum fluctuations into the conventional EM algorithm because quantum fluctuations induce the tunnel effect and are expected to relax the difficulty of nonconvex optimization problems in the maximum likelihood estimation problems.
We show a theorem that guarantees its convergence and give numerical experiments to verify its efficiency.
\end{abstract}

%% file: paper-acc-2016-03-01-intro.tex
\section{Introduction} \label{intro}

Many of practical problems in engineering or the principles to explain the phenomena of nature are reduced into nonconvex optimization; however, the research activity for nonconvex optimization is limited compared to that on convex optimization because it is fundamentally difficult to solve or analyze the problem in a sophisticated way.

In order to solve this difficulty of nonconvex optimization, motivated by the physical process of annealing, Kirkpatrick \textit{et al.}~\cite{Kirkpatrick01, Kirkpatrick02} proposed simulated annealing (SA).
SA has attracted much attention in many fields because SA has the following two remarkable properties.
The first one is that SA can be applied to any nonconvex problems.
The second one is that its global convergence in some sense is guaranteed by Geman and Geman~\cite{Geman01}.
After that, quantum annealing (QA) was proposed by Apolloni \textit{et al.}~\cite{Apolloni01}.
In QA, the mathematical mechanism of quantum fluctuations is introduced, and it has been reported that QA can reduce computational costs in many difficult problems~\cite{Finnila01, Kadowaki01, Brooke01, Falco01, Falco02, Das01, Farhi01, Santoro01, Santoro02, Marto01}.
Despite the success of SA and QA, their computational costs are still huge, because the Monte Carlo method is used in most of their implementations and it requires much computational costs for convergence.

On the other hand, in order to solve the nonconvex problem in data clustering, Rose \textit{et al.}~\cite{Rose01, Rose02} proposed a deterministic simulated annealing approach, and it attracted interest in both of physics and engineering.
This is because it can relax the problem of local optima with almost the same numerical cost which is required for a conventional approach.
As a generalization of~\cite{Rose01, Rose02}, Ueda and Nakano~\cite{Ueda01} proposed a deterministic simulated annealing EM algorithm (DSAEM).
The EM algorithm (EM), which was originally proposed by Dempster \textit{et al.}~\cite{Dempster01}, is a generic approach to compute the maximum likelihood estimates, but it is known that it suffers from the problem of local optima.
DSAEM was reported to be more effective than EM when EM is likely to be trapped by local optima; however the problem of local optima is still fundamental in optimization and it has been tackled via many approaches.

From the above discussions, this paper presents a deterministic quantum annealing EM algorithm (DQAEM) by introducing the mathematical mechanism of quantum fluctuations to EM.
In DQAEM, quantum fluctuations are introduced because it may induce the tunnel effect and it is considered to be effective to solve nonconvex optimization problems.
However, it is known to be difficult to evaluate functions with quantum fluctuations, and our key idea to compromise this difficulty is to apply the Feynman path integral formulation to DQAEM.
In this paper, after explaining EM, we give the formulation of DQAEM and how it is approximated through the Feynman path integral formulation.
Then, we present a theorem that ensures the monotonicity of its cost function, called the ``free energy," during the iterations in the algorithm.
That is, DQAEM is guaranteed to converge to the global optimum or local optima.
At the end of this paper, in order to show the efficiency of DQAEM, we apply DQAEM and EM to a parameter estimation problem and illustrate that DQAEM is superior to EM.
We summarize the algorithms mentioned above in Table~\ref{table01} for convenience.

This paper is organized as follows.
In Sec.~\ref{review-em}, EM proposed by Dempster \textit{et al.}~\cite{Dempster01} is reviewed for preparation of DQAEM.
In Sec.~\ref{qaem}, we propose DQAEM and give a theorem on its convergence.
In Sec.~\ref{sec-numerical}, we show numerical simulations to verify the theorem and the efficiency of DQAEM compared with EM.
Finally, in Sec.~\ref{conc}, we give the conclusion of this paper.

\begin{table}[t] 
\caption{Classification of the algorithms}
\label{table01}
\begin{center}
\begin{tabular}{|c|c|c|}
\hline
Fluctuations &
Annealing~\footnote{Conventional stochastic annealing for optimization} &
DAEM~\footnote{Deterministic annealing EM algorithm (DAEM)} \\
\hline
Thermal & Kirkpatrick \textit{et al.}~\cite{Kirkpatrick01,Kirkpatrick02}~\footnote{Simulated annealing (SA)} & Ueda \& Nakano~\cite{Ueda01}~\footnote{Deterministic simulated annealing EM algorithm (DSAEM)} \\
Quantum & Apolloni \textit{et al.}~\cite{Apolloni01}~\footnote{Quantum annealing (QA)} & This work~\footnote{Deterministic quantum annealing EM algorithm (DQAEM)} \\
\hline
\end{tabular}
\end{center}
\end{table}

%% file: paper-acc-2016-05-00-em.tex
\section{Review of the EM algorithm (EM)} \label{review-em}


%% file: paper-acc-2016-05-01-em.tex

In this section, we review EM, which was proposed by Dempster \textit{et al.}~\cite{Dempster01} for the preparation of introducing DQAEM.
In this paper, $Y_\mathrm{obs} = \{y^{(1)}, y^{(2)}, \dots, y^{(N)} \}$ and $\{x^{(1)}, x^{(2)}, \dots, x^{(N)} \}$ denote an observable data set and an unobservable data set, respectively.
Moreover, we assume that each data point $y^{(i)} \ (i=1, 2, \dots, N)$ is independent and identically distributed.
Then, at first, we define maximum likelihood estimation (MLE).
In MLE, the cost function, which is called the log likelihood function, is given by
\begin{align}
\mathcal{L}(Y_\mathrm{obs};\theta) &= \sum_{i=1}^N \log p(y^{(i)};\theta) \nonumber \\
                      &= \sum_{i=1}^N \log \int dx^{(i)}\, p(y^{(i)}, x^{(i)};\theta), \label{log-likelihood-05}
\end{align}
where $p(y; \theta)$ and $p(y, x; \theta)$ are probability density functions for incomplete data and complete data, respectively, and $\theta$ is a parameter.
Then the parameter $\theta$ is determined by the maximization of $\mathcal{L}(Y_\mathrm{obs};\theta)$ with respect to $\theta$.
Note that in the case that $p(y; \theta)$ belongs to the exponential family, the maximization can be easily attained.

In most practical cases, however, $p(y; \theta)$ does not belong to the exponential family, and then the maximization of $\mathcal{L}(Y_\mathrm{obs}; \theta)$ is difficult to compute.
EM was proposed as an iterative approach to calculate the maximum likelihood estimates in these cases, and then the maximization of $\mathcal{L}(Y_\mathrm{obs}; \theta)$ is replaced by its lower bound called the $Q$ function.
We then derive the $Q$ function as follows:
\begin{align}
\mathcal{L}(Y_\mathrm{obs};\theta) &\ge Q(\theta;\theta') \nonumber \\
                      & \quad - \sum_{i=1}^N \int dx^{(i)}\, p(x^{(i)}|y^{(i)};\theta') \log p(x^{(i)}|y^{(i)};\theta'), \label{free-Q-11} \\
Q(\theta;\theta')     &= \sum_{i=1}^N \int dx^{(i)}\, p(x^{(i)}|y^{(i)};\theta') \log p(y^{(i)},x^{(i)};\theta), \label{EM-Mstep01}
\end{align}
where the parameter $\theta'$ is an arbitrary parameter and Jensen's inequality is used to derive the inequality.
Note that the $Q$ function includes a conditional probability density function $p(x| y; \theta)$, and it is computed with Bayes' rule as
\begin{align}
p(x^{(i)}|y^{(i)};\theta') = \frac{p(y^{(i)},x^{(i)};\theta')}{\int dx\, p(y^{(i)},x;\theta')}, \label{EM-Estep01}
\end{align}
for $i=1, 2, \dots, N$.
Then, the parameter $\theta^{(t)}$ is updated by $\theta' = \theta^{(t)}$ and
\begin{align}
\theta^{(t+1)} = \argmax_\theta Q(\theta; \theta^{(t)}). \label{parameter-update-01}
\end{align}
The calculation of~\eqref{EM-Mstep01} is called the E step and that of~\eqref{parameter-update-01} is called the M step, and they are iterated until termination conditions are satisfied.
We summarize EM in Table~\ref{EM-algorithm01}.
\begin{algorithm}[t]
\caption{The EM algorithm (EM)}
\label{EM-algorithm01}
\begin{algorithmic}[1]
\STATE initialize $\theta^{(0)}$ and set $t \leftarrow 0$
\WHILE{convergence criterion is satisfied}
\STATE calculate $p(x^{(i)}|y^{(i)};\theta^{(t)})$ for $i \ (i=1, 2, \dots, N)$ with~\eqref{EM-Estep01} \ (E step)
\STATE calculate $\theta^{(t+1)} = \argmax_\theta Q(\theta;\theta^{(t)})$ where $Q(\theta;\theta^{(t)})$ is~\eqref{EM-Mstep01} \ (M step)
\ENDWHILE
\end{algorithmic}
\end{algorithm}

EM is widely used because it has two remarkable properties as follows.
The first one is that it can be applied to mixture models, which often appear in practical applications, and it gives better performance than conventional methods in many cases.
The second one is that the log likelihood function monotonically increase via its iterations, and this implies that it is stable through iterations.
However it is also known to be trapped by local optima and its performance heavily depends on an initial estimated parameter, and thus this problem is a motivation of our study.

%% file: paper-acc-2016-06-00-dqaem.tex
\section{Deterministic quantum annealing EM algorithm (DQAEM)} \label{qaem}

Here, we derive DQAEM, which is the main concept of this paper, by introducing the mechanism of quantum fluctuations into EM.
We then give a theorem which guarantees its convergence.

%% file: paper-acc-2016-06-01-dqaem.tex
\subsection{Derivation} \label{qaem-sub-01}

First, we define the cost function, which is called the free energy, as
\begin{align}
F_{\beta,\Gamma}(\theta) &= -\frac{1}{\beta} \log \mathcal{Z}_{\beta,\Gamma} (\theta), \label{free-dqa-04}
\end{align}
where
\begin{align}
\mathcal{Z}_{\beta, \Gamma} (\theta) &= \prod_{i=1}^N \mathcal{Z}_{\beta, \Gamma}^{(i)} (\theta), \label{partition-94} \\
\mathcal{Z}_{\beta, \Gamma}^{(i)} (\theta) &= \int dx^{(i)}\, \langle x^{(i)} | p_\Gamma(y^{(i)}, \hat{x}; \theta)^\beta | x^{(i)} \rangle, \nonumber \\
p_\Gamma(y^{(i)}, \hat{x}; \theta) &= \exp \{- (H(y^{(i)}, \hat{x}; \theta) + H_\mathrm{kin})\} \ (i = 1, 2, \dots, N), \label{quantum-dist-77}
\end{align}
$\hat{x}$ represents a position operator, $H(y^{(i)}, \hat{x}; \theta) = - \log p(y^{(i)}, \hat{x}; \theta)$, and $H_\mathrm{kin}$ is a function of a momentum operator $\hat{\pi}$.
This momentum operator $\hat{\pi}$ satisfies the commutation relation $[\hat{x}, \hat{\pi}] = i \hbar$.
In this paper, we consider the form $H_\mathrm{kin} = \hat{\pi}^2 / 2 \mu$ to simplify later calculations.
From the definition of the free energy~\eqref{free-dqa-04} and the log likelihood function~\eqref{log-likelihood-05}, they hold the following identity,
\begin{align}
F_{\beta=1,\Gamma=0}(\theta) &= -\mathcal{L}(Y_\mathrm{obs}; \theta). \label{equality-QA-01}
\end{align}
We then interpret the negative free energy as an extension of the log likelihood function.

Second, in order to formulate DQAEM, we divide the free energy~\eqref{free-dqa-04} into two parts as follows,
\begin{align}
F_{\beta,\Gamma}(\theta) &= U_{\beta,\Gamma}(\theta; \theta') - \frac{1}{\beta} S_{\beta,\Gamma}(\theta; \theta'), \nonumber \\
U_{\beta,\Gamma} &(\theta; \theta') \nonumber \\
                 &= \sum_{i=1}^N \int dx^{(i)}\, \left\langle x^{(i)} \middle| \left[ -f_{\beta, \Gamma}(\hat{x}^{(i)}| y^{(i)}; \theta') \log p_\Gamma (y^{(i)}, \hat{x}^{(i)}; \theta) \right] \middle| x^{(i)} \right\rangle, \label{QAEM-Mstep01} \\
S_{\beta,\Gamma} &(\theta; \theta') \nonumber \\
                 &= \sum_{i=1}^N \int dx^{(i)}\, \left\langle x^{(i)} \middle| \left[ -f_{\beta, \Gamma}(\hat{x}^{(i)}| y^{(i)}; \theta') \log f_{\beta, \Gamma}(\hat{x}^{(i)}| y^{(i)}, \theta) \right] \middle| x^{(i)} \right\rangle, \label{QAEM-entropy01}
\end{align}
with the operator of the perturbed conditional probability density function,
\begin{align}
f_{\beta, \Gamma} (\hat{x}^{(i)}|y^{(i)}; \theta) = \frac{p_\Gamma (y^{(i)}, \hat{x}^{(i)}; \theta)^\beta}{\mathcal{Z}_{\beta, \Gamma}^{(i)} (\theta)}. \label{posterior02}
\end{align}
The function $U_{\beta, \Gamma}(\theta, \theta')$ is an extension of the $Q$ function~\eqref{EM-Mstep01}, and satisfies the identity
\begin{align}
U_{\beta=1, \Gamma=0}(\theta, \theta') = - Q(\theta, \theta'). \nonumber
\end{align}
As the $Q$ function is optimized instead of the log likelihood function in EM, the function $U_{\beta, \Gamma}(\theta, \theta')$ is optimized in DQAEM.

In some special case, the calculations of the free energy~\eqref{free-dqa-04} and the energy~\eqref{QAEM-Mstep01} can be performed analytically; however those are difficult in general when an assumed model has a complicated form.
We thus use the Feynman's path integral formula~\cite{Feynman01, Feynman02, Takahashi01, Takahashi02} to simplify the calculation of $U_\mathrm{\beta, \Gamma}$ as
\begin{align}
&U_{\beta, \Gamma} (\theta; \theta') \nonumber \\
& \quad \approx \sum_{i=1}^N \int \prod_{j=1}^M dx_j^{(i)}\, \Bigg[ \frac{-1}{M}  f_{\beta, \Gamma}(\{x_j^{(i)}\}| y^{(i)}; \theta') \sum_{j=1}^M \log p(y^{(i)},x_j^{(i)};\theta) \Bigg] \nonumber \\
& \qquad + \mathrm{const.}, \label{QAEM-Mstep02}
\end{align}
with
\begin{align}
f_{\beta, \Gamma}(\{x_j^{(i)}\}| y^{(i)}; \theta') & = \frac{1}{\mathcal{Z}_{\beta, \Gamma}^{(i)} (\theta')} \left( \frac{M}{2 \pi \beta \Gamma} \right)^{M/2} \nonumber \\
                                       & \quad \exp \Bigg( \sum_{j=1}^M \frac{\beta}{M} \log p(y^{(i)},x_j^{(i)};\theta') \nonumber \\
                                       & \qquad - \sum_{j=1}^M \frac{M}{2 \beta \Gamma} (x_j^{(i)} - x_{j-1}^{(i)})^2 \Bigg),\label{QAEM-Estep04}
\end{align}
where $M$ is the number of beads, $\{x_j^{(i)}\}$ represents $\{x_j^{(i)}\}_{j=1}^M$ and the periodic boundary conditions $x_0^{(i)} = x_M^{(i)}$ for each $i \ (i=1, 2, \dots, N)$ are satisfied.
The updating equation for the parameter $\theta$ is therefore given by
\begin{align}
\theta^{(t+1)} = \argmin_\theta U_{\beta,\Gamma}(\theta; \theta^{(t)}). \nonumber
\end{align}

We put the above calculations to an algorithm as DQAEM, which includes two steps called the E step and the M step.
In the E step of DQAEM, $f_{\beta, \Gamma} (\{x_j^{(i)}\}| y^{(i)}; \theta^{(t)})$ in~\eqref{QAEM-Estep04} is computed, and in the M step of DQAEM, the parameter $\theta^{(t)}$ is updated by the minimization of~\eqref{QAEM-Mstep02}.
Finally, we summarize the algorithm in Table~\ref{QAEM-algorithm-01}.
\begin{algorithm}[t]
\caption{Deterministic quantum annealing EM algorithm (DQAEM)}
\label{QAEM-algorithm-01}
\begin{algorithmic}[1]
\STATE set $\beta \leftarrow \beta_\mathrm{init} (0 < \beta_\mathrm{init} \le 1)$
\STATE set $\Gamma \leftarrow \Gamma_\mathrm{init} (0 \le \Gamma_\mathrm{init})$
\STATE initialize $\theta^{(0)}$ and set $t \leftarrow 0$
\WHILE{convergence criteria is satisfied}
\STATE calculate $f_{\beta, \Gamma} (\{x_j^{(i)}\}| y^{(i)}; \theta^{(t)})$ for $i \ (i=1, 2,\dots,N)$ with~\eqref{QAEM-Estep04} \ (E step)
\STATE calculate $\theta^{(t+1)} = \argmin_{\theta} U_{\beta,\Gamma}(\theta;\theta^{(t)})$ with~\eqref{QAEM-Mstep02} \ (M step)
\STATE increase $\beta$ and decrease $\Gamma$
\ENDWHILE
\end{algorithmic}
\end{algorithm}

%% file: paper-acc-2016-06-02-dqaem.tex
\subsection{Convergence theorem} \label{qaem-sub-02}

In general, stability of an algorithm is not obvious, and here we give a theorem that guarantees the stability of DQAEM through iterations as follows.
\begin{theorem} \label{theorem01}
Let the parameter $\theta^{(t+1)}$ be given by $\theta^{(t+1)} = \argmin_\theta U_{\beta, \Gamma}(\theta;\theta^{(t)})$.
Then the inequality $F_{\beta, \Gamma}(\theta^{(t+1)}) \le F_{\beta, \Gamma}(\theta^{(t)})$ holds.
The equality holds if and only if $U_{\beta, \Gamma}(\theta^{(t+1)}; \theta^{(t)}) = U_{\beta, \Gamma}(\theta^{(t)};\theta^{(t)})$ and $S_{\beta, \Gamma}(\theta^{(t+1)};\theta^{(t)}) = S_{\beta, \Gamma}(\theta^{(t)};\theta^{(t)})$ are satisfied.
\end{theorem}
By this theorem, we can conclude that DQAEM is guaranteed to converge to the global optimum or a local optimum.
It is known that the monotonicity of the log likelihood function in EM and some mathematical features of EM are proved by Dempster \textit{et al.}~\cite{Dempster01} and Wu~\cite{Wu01}, and this theorem clarifies that the similar monotonicity also holds in DQAEM.

%% file: paper-acc-2016-07-00-result.tex
\section{Numerical simulations} \label{sec-numerical}

In this section, we give numerical simulations to show the efficiency of DQAEM in comparison with EM.
At, first we begin with the definition of the problem that we consider in this section.
Next, we give the numerical simulations which support the theorem shown in the previous section.
Finally, we compare DQAEM and EM by applying them to the problem and discuss the efficiency of DQAEM.


%% file: paper-acc-2016-07-01-result.tex
\subsection{Mathematical formulation} \label{sec-numerical-01}

We adopt the mixture of factor analysis (MFA)~\cite{Ghahramani04, Murphy01} as the problem to which DQAEM and EM are applied in this section.
MFA is a model to analyze hidden factors in a given data set, and is a typical nonconvex optimization problem if the number of factors is larger than $1$.
Accordingly, EM is often applied to MFA for practical calculations.

MFA can be considered to assume following two steps to generate data.
In the first step, a factor is identified by an index parameter $w \in \{1, 2, \dots, m\}$, which is generated with a probability $P(w)$, and an unobservable state $x$ is generated with a probability density function $p(x)$.
In the second step, the observable variable $y$ is generated by the transformation of $x$ depending on $w$ and an additive noise.
This transformation is represented by $p(y| x, w; \theta)$.
The probability density function for $y$ is then given by
\begin{align}
p(y; \theta) &= \sum_{w=1}^m \int dx\, p(y, x, w; \theta), \label{MFA-eq-01} \\
p(y, x, w; \theta) &=  p(y| x, w; \theta) p(x) P(w), \label{MFA-model-01}
\end{align}
where $P(w = i) = \pi_i$, $p(x) = \mathcal{N}(0, I)$, $p(y| x, w = i; \theta) = \mathcal{N} (\mu_i + \Lambda_i x, \Phi)$, and $m$ is the assumed number of factors in MFA.
For simplicity, we denote $\{ \pi_i, \mu_i, \Lambda_i, \Phi\}_{i=1}^m$ by $\theta$.

In DQAEM for MFA, the Hamiltonian $H_\mathrm{kin}$, which represents the kinetic term, is added to the original model of MFA, and then we have the following
\begin{align}
p_\Gamma(y, \hat{x}, w; \theta) &=  p(y| \hat{x}, w; \theta) e^{- (H + H_\mathrm{kin})} P(w), \label{MFA-model-02}
\end{align}
where $H = - \log p(\hat{x})$ and $H_\mathrm{kin} = \hat{\pi}^2 / 2 \mu$ ($H_\mathrm{kin} = - \Gamma \partial^2 / \partial x^2$ with $\Gamma = \hbar^2 / \mu$ in $x$-bases).
Suppose we have $N$ data points $Y_\mathrm{obs} = \{y^{(1)}, y^{(2)}, \dots, y^{(N)}\}$, and then the free energy is given by
\begin{align}
F_{\beta, \Gamma} (\theta) &= \sum_{i=1}^N - \frac{1}{\beta} \log \sum_{w^{(i)}=1}^m \int dx^{(i)}\, \left< x^{(i)} \middle| p_\Gamma(y^{(i)}, \hat{x}^{(i)}, w^{(i)}; \theta)^\beta \middle| x^{(i)} \right>. \label{MFA-eq-02}
\end{align}
In numerical experiments, $M$ is set to 128, the annealing parameter $\Gamma$, which represents the strength of quantum fluctuations, is controlled from an initial value to $0$ linearly, and the inverse temperature $\beta$ is fixed at $1$.

%% file: paper-acc-2016-07-02-result.tex
\subsection{Numerical results I} \label{sec-numerical-02}

We have shown the monotonicity of the free energy~\eqref{free-dqa-04} in DQAEM in Sec.~\ref{qaem-sub-02}, and here we verify Theorem~\ref{theorem01} via numerical experiments.
In this subsection, we consider 4 models which have 1, 3 ,7 and 10 mixtures, respectively.

\begin{figure}[t]
\begin{subfigure}[t]{0.5\textwidth}
\centering
\includegraphics[scale=0.50]{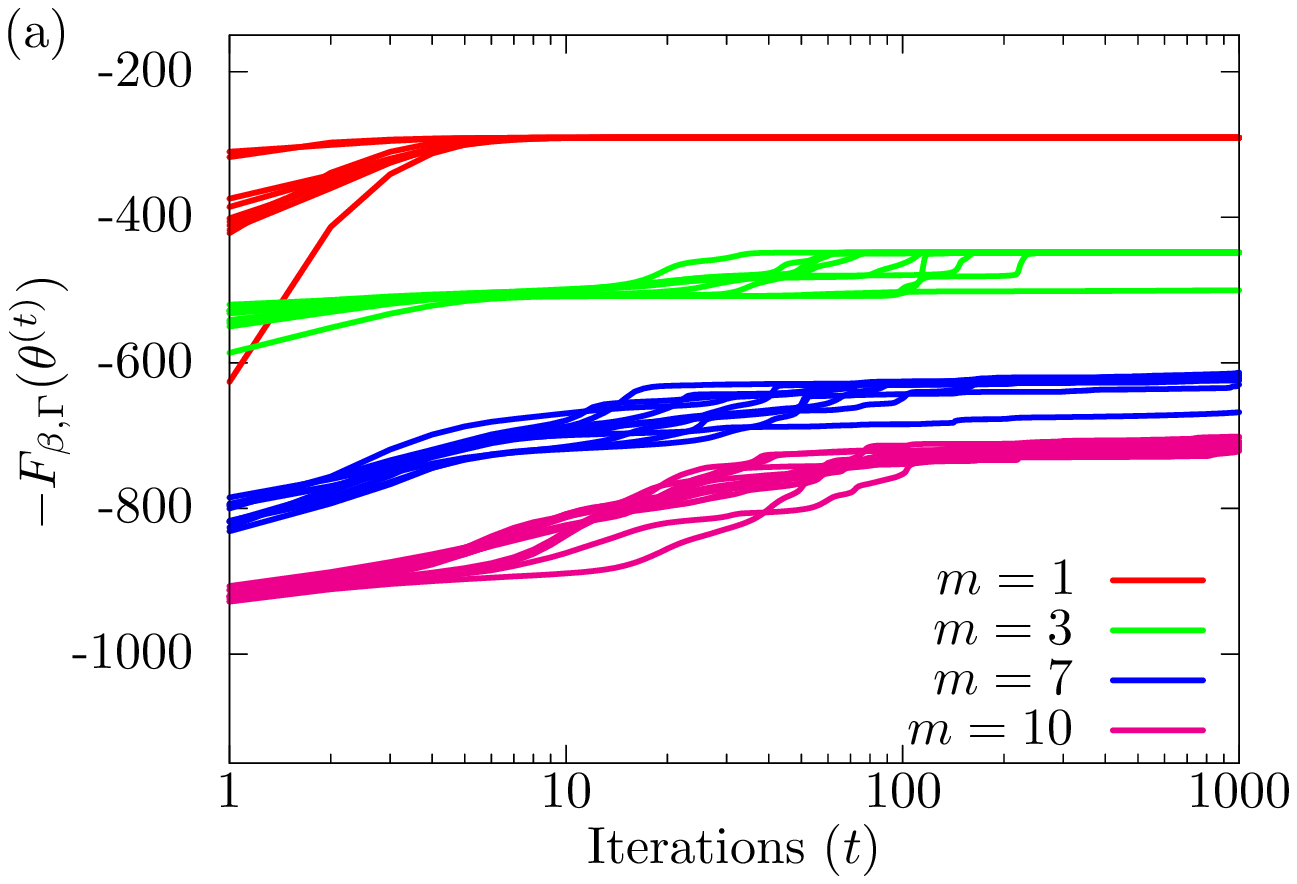}
\phantomcaption
\label{numerical-sub-01-01}
\end{subfigure}
\begin{subfigure}[t]{0.5\textwidth}
\centering
\includegraphics[scale=0.50]{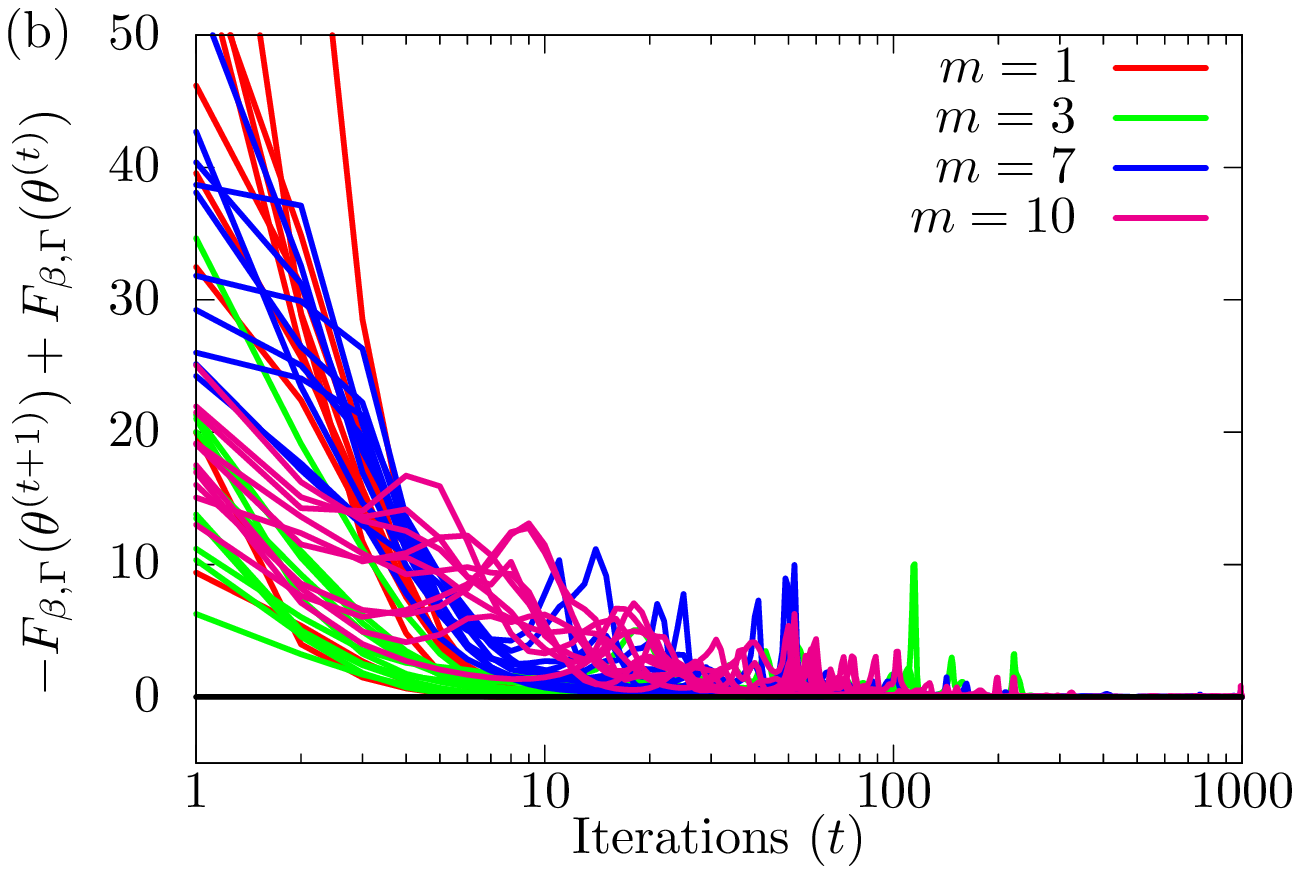}
\end{subfigure}
\caption{Number of iterations vs (a) negative free energy and (b) difference of negative free energies in each step for 4 models whose numbers of mixtures are $m=1, 3, 7$ and $10$.}
\label{numerical-03-84}
\end{figure}

The transitions of negative free energies $-F_{\beta, \Gamma} (\theta^{(t)})$ for these models are shown in Fig.~\ref{numerical-03-84}(a), and the change of them $-F_{\beta, \Gamma} (\theta^{(t+1)}) + F_{\beta, \Gamma} (\theta^{(t)})$ is plotted in Fig.~\ref{numerical-03-84}(b).
By observing Fig.~\ref{numerical-03-84}, we can confirm that the negative free energy varies monotonically in DQAEM.
In the case that $m = 1$, this problem is a convex optimization, and thus we can see that the negative free energies converge to the unique optimal value in Fig.~\ref{numerical-03-84}(a).

%% file: paper-acc-2016-07-03-result.tex
\subsection{Numerical results II} \label{sec-numerical-03}

\begin{figure}[t]
\begin{subfigure}[t]{0.5\textwidth}
\centering
\includegraphics[scale=0.50]{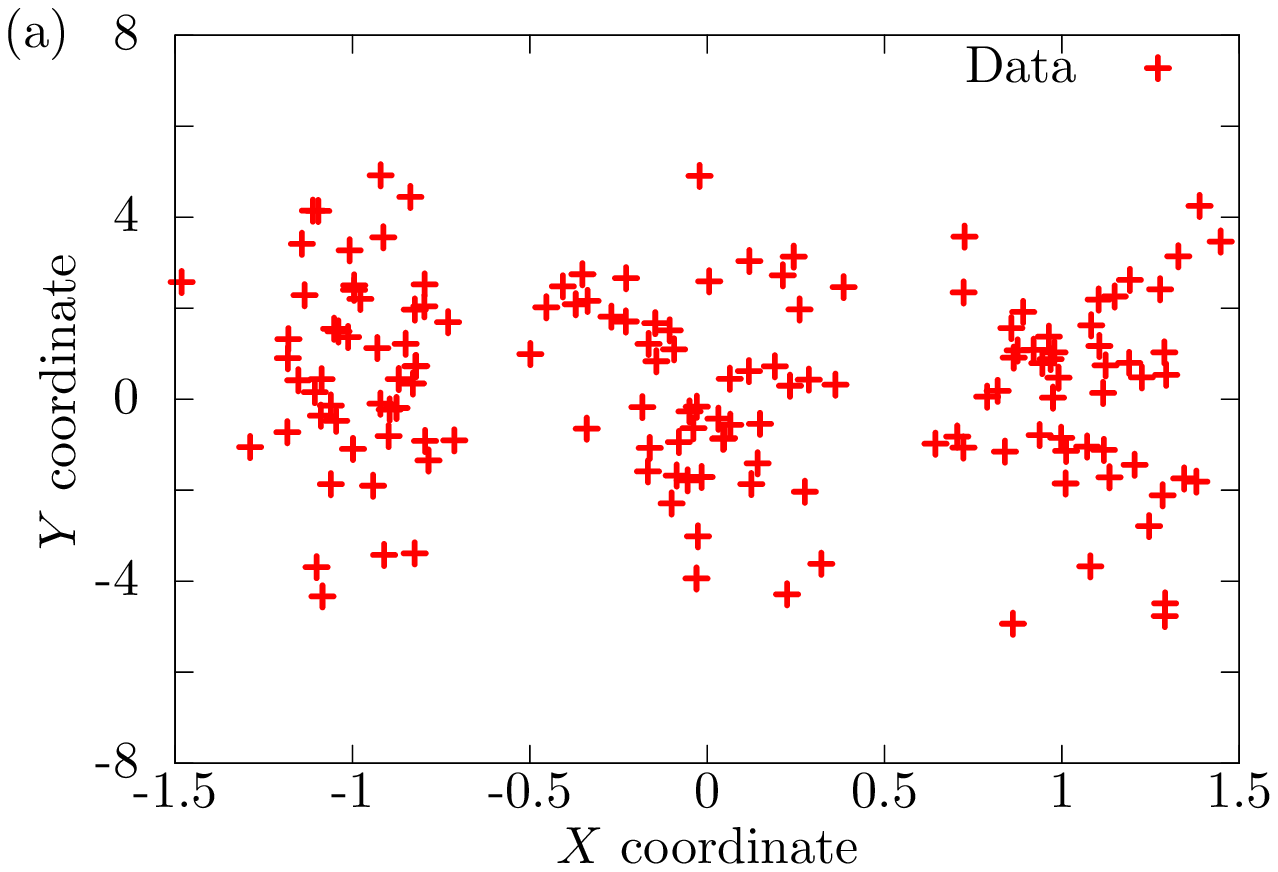}
\end{subfigure}
\begin{subfigure}[t]{0.5\textwidth}
\centering
\includegraphics[scale=0.50]{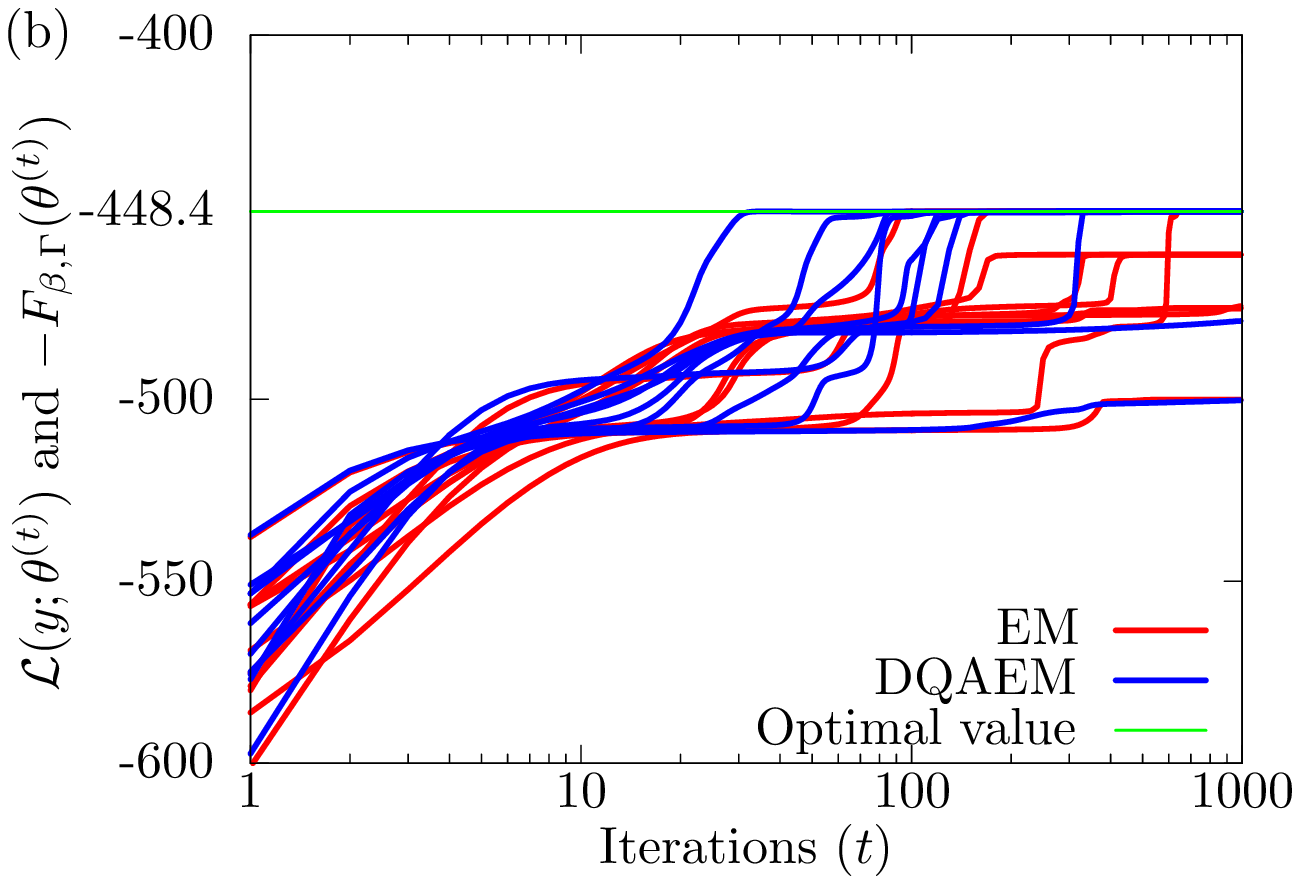}
\end{subfigure}
\caption{(a) Data set generated by three Gaussian functions whose means are $(X, Y) = (-1, 0)$, $(0, 0)$ and $(1,0)$. (b) Number of iterations (log scale) vs the log likelihood functions in EM and the negative free energies in DQAEM.}
\label{numerical-01-01}
\end{figure}

This subsection is devoted to the comparison between DQAEM and EM to show the efficiency of DQAEM.
We deal with the data shown in Fig.~\ref{numerical-01-01}, which are generated by the Gaussian mixture model whose means are $(X, Y) = (-1, 0)$, $(0, 0)$ and $(1,0)$.

In Fig.~\ref{numerical-01-01}(b), the transients of the log likelihood functions by EM are plotted by red lines and those of the negative free energies by DQAEM are plotted by blue lines.
The green line represents the value of $-448.4$, which corresponds to the optimal estimation in this problem.
Some of red lines and blue lines converge to the value of $-448.4$, and thus DQAEM and EM give the optimal estimation in these cases.
On the other hand, some of red lines and blue lines converge to lower values than the optimal estimation.

However, the ratios whether DQAEM and EM gives the optimal estimation are much different.
We performed DQAEM and EM 1000 times with the same initial conditions, and the ratios whether DQAEM and EM succeed or fail with the same randomized initial parameters are summarized in Table~\ref{joint-table-01}.
\begin{table}[t]
\caption{Ratios whether DQAEM and EM succeed or fail for this problem (The details of the cases labeled by $^{(*)}$ and $^{(**)}$ are discussed in Secs.~\ref{sec-numerical-03-01} (Case I$^{(*)}$) and \ref{sec-numerical-03-02} (Case II$^{(**)}$), respectively).}
\label{joint-table-01}
\begin{center}
\begin{tabular}{|c|c|p{5em}|p{5em}|p{5em}|}
\hline
\multicolumn{2}{|c|}{}  & \multicolumn{3}{c|}{DQAEM}   \\ \cline{3-5}
\multicolumn{2}{|c|}{}  & Success & Fail    & Total      \\ \hline
                        & Success     & 36.3 \%$^{(*)}$ &  3.3 \% & 36.6  \% \\ \cline{2-5}
EM        & Fail        & 54.4 \%$^{(**)}$ & 9.0 \% & 63.4  \% \\ \cline{2-5}
          & Total       & 90.7 \% & 9.3 \% & 100.0 \% \\ \hline
\end{tabular}
\end{center}
\end{table}
Here, the ``success" of DQAEM and EM is defined as that square errors between the estimated means of three Gaussian functions and the true means are smaller than $0.2$ times the covariances of three Gaussian functions.
This table shows that DQAEM succeeds with the ratio 90.7\% while EM succeeds with the ratio 36.6\%, and that DQAEM is superior to EM.

%% file: paper-acc-2016-07-04-result.tex
\subsubsection{Case I$^{(*)}$} \label{sec-numerical-03-01}

Here we focus on the convergence rates of DQAEM and EM in the case that both of them succeed in parameter estimation.

\begin{figure}[t]
\begin{subfigure}[t]{0.50\textwidth}
\centering
\includegraphics[scale=0.50]{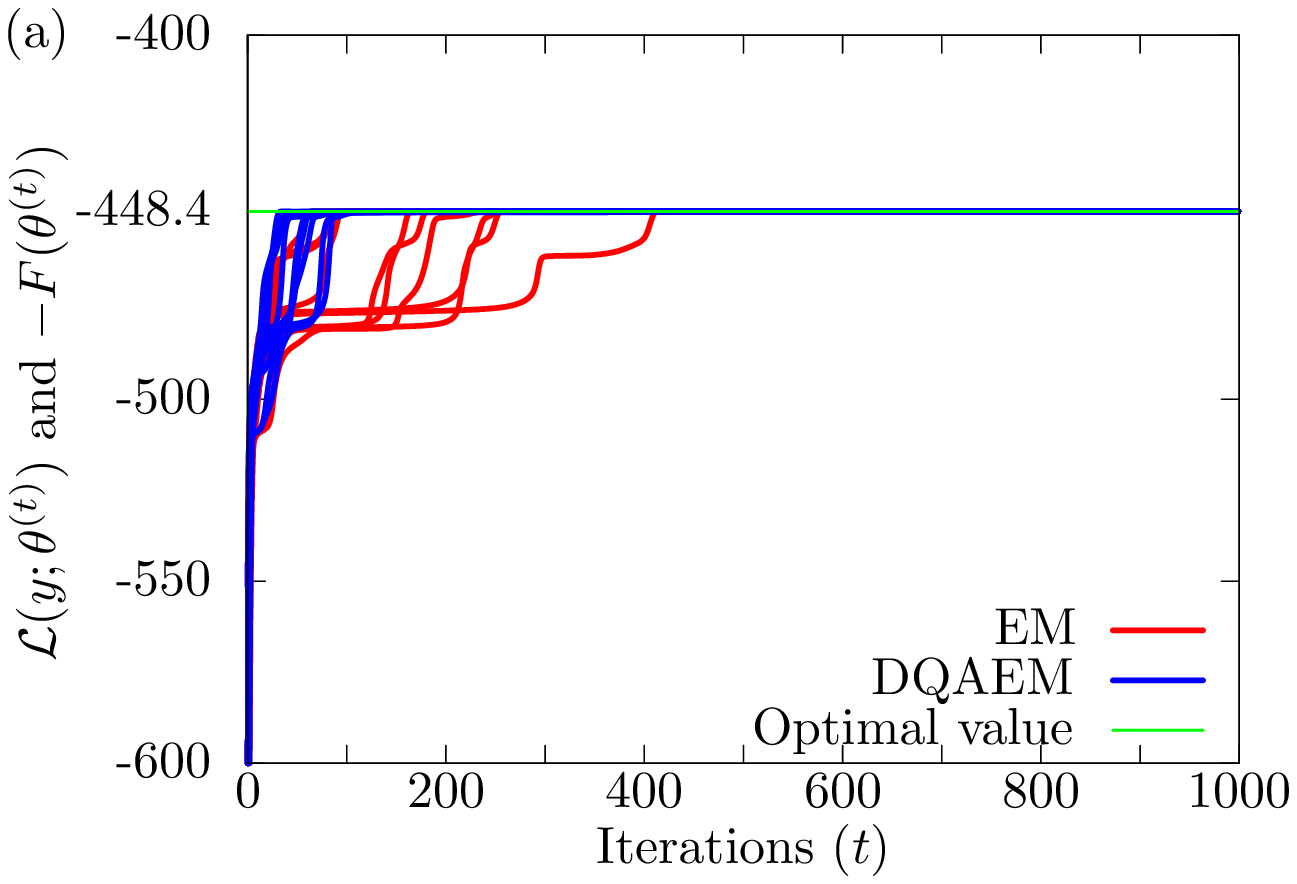}
\end{subfigure}
\begin{subfigure}[t]{0.50\textwidth}
\centering
\includegraphics[scale=0.50]{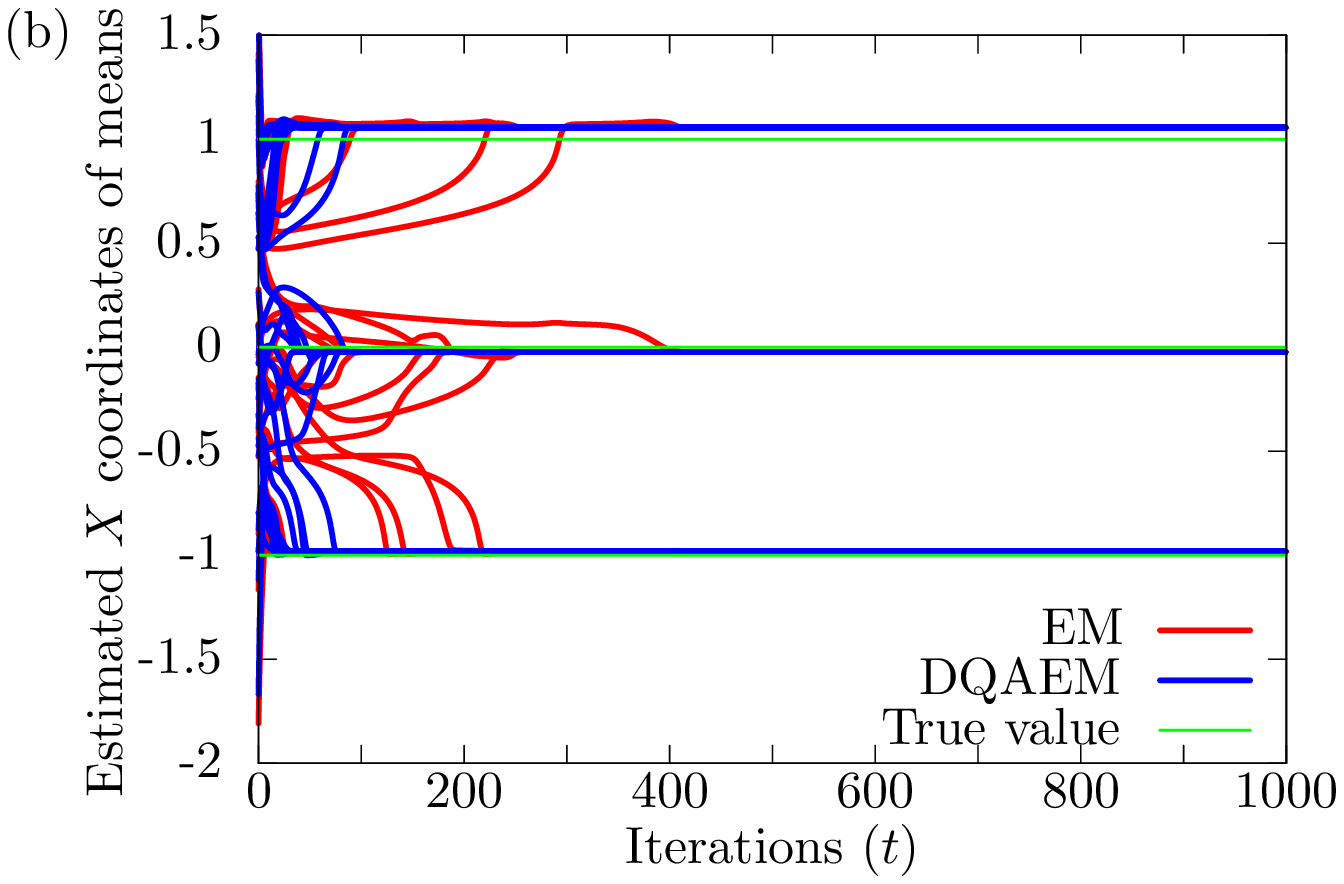}
\end{subfigure}
\caption{(a) Number of iterations vs the log likelihood functions of EM and the negative free energies of DQAEM with the initial estimated parameters with which both EM and DQAEM give global optimums (Case I$^{(*)}$). Green line exhibits the optimal value. (b) Number of iterations vs the estimated $X$ coordinates of means of Gaussian functions by EM, and those by DQAEM in Case I$^{(*)}$. Green lines stand for the true values.}
\label{numerical-02-01}
\end{figure}

First, we plot the log likelihood functions and the negative free energies in Fig.~\ref{numerical-02-01}(a) and the estimated $X$ coordinates in Fig.~\ref{numerical-02-01}(b).
In this case, both the log likelihood functions and the negative free energies converge to the value of $-448.4$, and the estimated $X$ coordinates go to the neighbors around $-1.0$, $0.0$ and $1.0$.
These figures also imply that DQAEM converges to the optimal estimation faster than EM.

We also show the number of iterations that are required until DQAEM and EM satisfy the criteria of the ``success" in Table~\ref{iteration-table-01}.
This table tells us that DQAEM is approximately 3.73 times faster than EM.
\begin{table}[t]
\caption{Numbers of iterations that are required for DQAEM and EM to satisfy the criteria of the ``success."}
\label{iteration-table-01}
\begin{center}
\begin{tabular}{|c|c|}
\hline
DQAEM        & EM           \\ \hline
65.33 times  & 243.82 times \\
\hline
\end{tabular}
\end{center}
\end{table}

%% file: paper-acc-2016-07-05-result.tex
\subsubsection{Case II$^{(**)}$} \label{sec-numerical-03-02}

\begin{figure}[t]
\begin{subfigure}[t]{0.50\textwidth}
\centering
\includegraphics[scale=0.50]{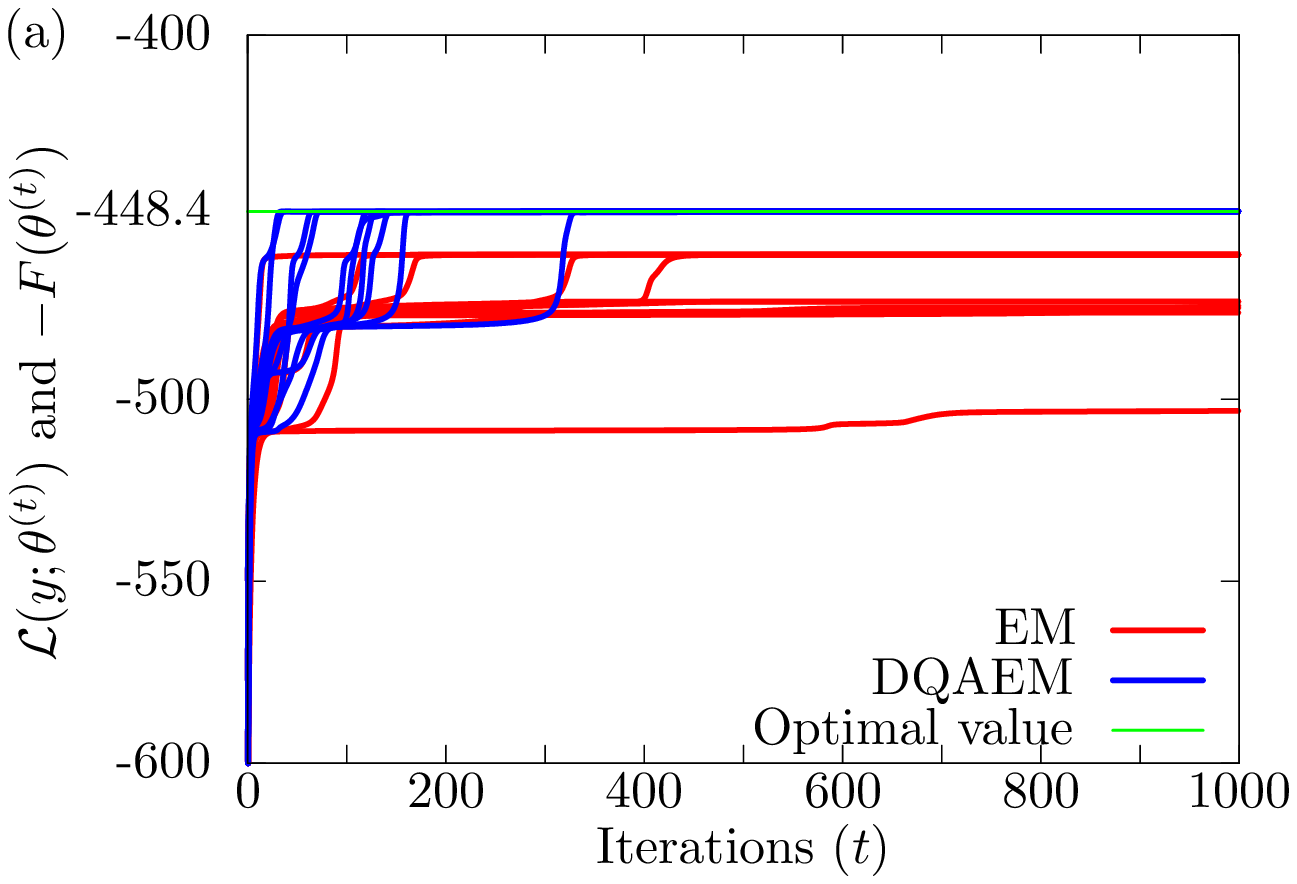}
\end{subfigure}
\begin{subfigure}[t]{0.50\textwidth}
\centering
\includegraphics[scale=0.50]{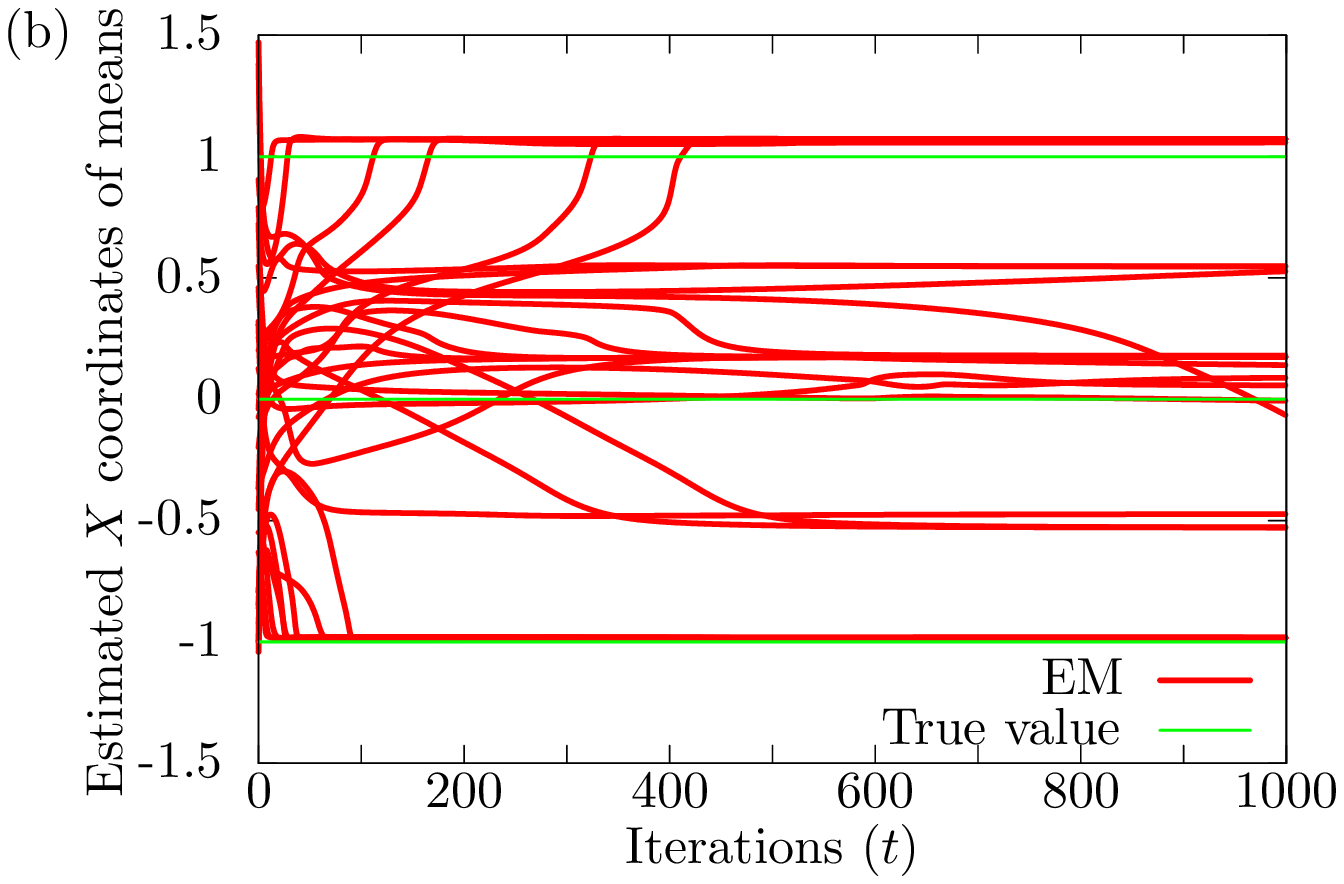}
\end{subfigure}
\begin{subfigure}[t]{0.50\textwidth}
\centering
\includegraphics[scale=0.50]{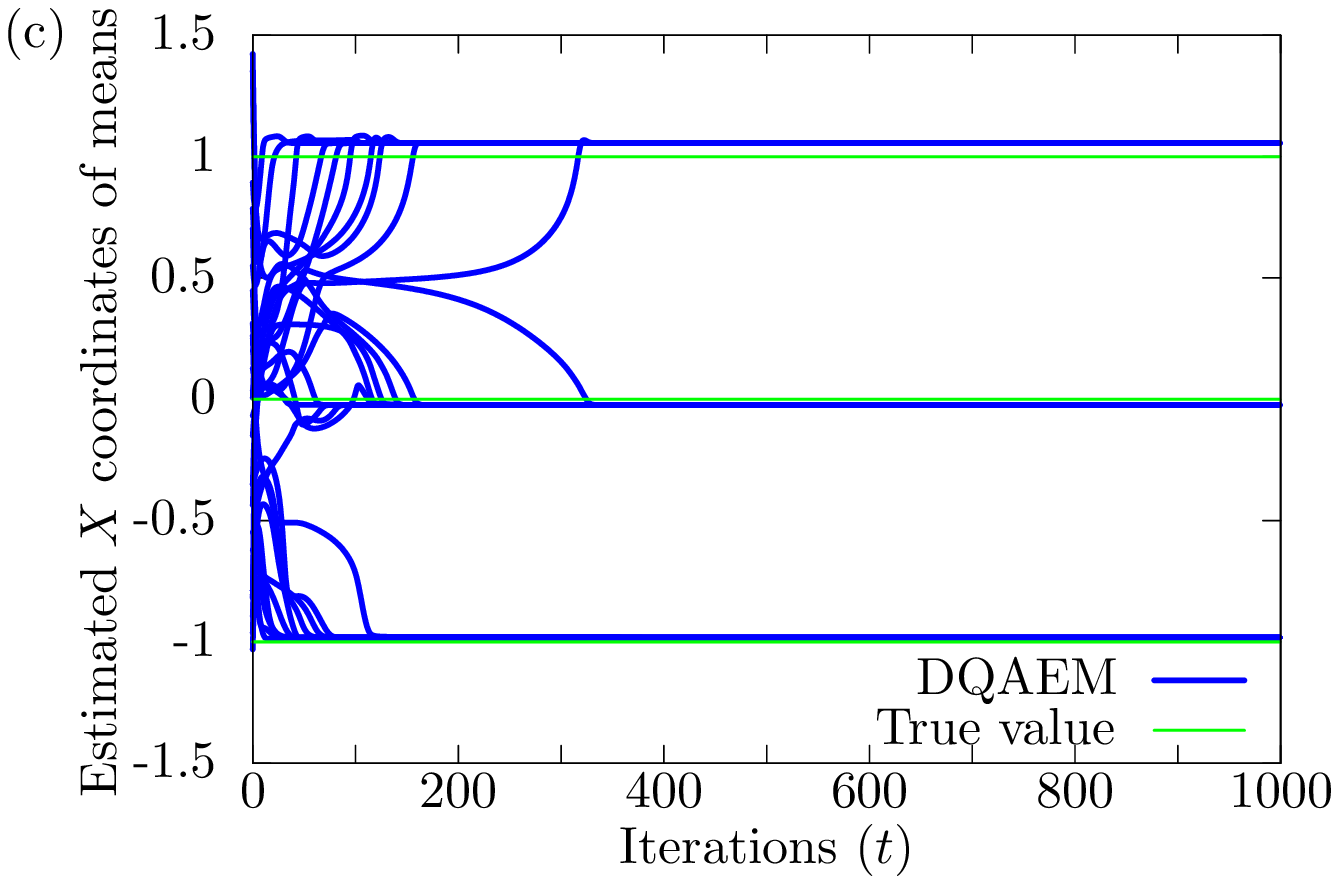}
\end{subfigure}
\caption{(a) Number of iterations vs the log likelihood functions of EM and the negative free energies of DQAEM with the initial estimated parameters with which EM fails and DQAEM succeeds (Case II$^{(**)}$). Green line stand for the optimal value. (b) Number of iterations vs the estimated $X$ coordinates of means of Gaussian functions by EM in Case II$^{(**)}$. Green lines stand for the true values. (c) Number of iterations vs the estimated $X$ coordinates of means of Gaussian functions by DQAEM in Case II$^{(**)}$. Green lines stand for the true values.}
\label{numerical-03-01}
\end{figure}

At the end of this section, we analyze the behaviors of DQAEM and EM in the case that DQAEM succeeds and EM fails.
In Fig.~\ref{numerical-03-01}(a), the log likelihood functions of EM and the negative free energies of DQAEM are plotted.
It is observed that while DQAEM gives the value of $-448.4$, EM gives lower values than that and is trapped by some local optima.
The estimated $X$ coordinates by EM and DQAEM are shown Figs.~\ref{numerical-03-01}(b) and (c), respectively.
While the estimated $X$ coordinates by DQAEM converge in the neighbors around $-1.0$, $0.0$ and $1.0$, the estimated $X$ coordinates by EM go to different values.

%% file: paper-acc-2016-08-02-conc.tex
\section{Conclusion} \label{conc}

We have proposed a deterministic quantum annealing EM algorithm (DQAEM) in this paper.
In DQAEM, the mathematical mechanism of quantum fluctuations is introduced into EM, and then our proposed algorithm is regarded as a quantum version of that by Ueda and Nakano~\cite{Ueda01}.
Then we show a theorem on the monotonicity of DQAEM mathematically, that is, it guarantees that DQAEM is stable in the algorithm iterations.
Through numerical experiments, we also confirmed the monotonicity of DQAEM, and then we compared DQAEM and EM.
In the comparison, we observe that DQAEM succeeds in parameter estimates with higher probability and faster than EM.
At the end, we mention our future work.
DQAEM in this paper is for models with continuous latent variables, and then one of our future works is to formulate DQAEM for models with discrete latent variables, which usually appear in various engineering problems.
